\documentclass[lettersize,journal]{IEEEtran}
\usepackage{amsmath,amsfonts}
\usepackage{algorithmic}
\usepackage{booktabs}
\usepackage{comment}
\usepackage{multirow}
\usepackage{algorithm}
\usepackage{array}
\usepackage{textcomp}
\usepackage{stfloats}
\usepackage{url}
\usepackage{verbatim}
\usepackage{times}
\usepackage{epsfig}
\usepackage{graphicx}
\usepackage{newtxmath}
\usepackage{xcolor}
\usepackage[pagebackref,breaklinks,colorlinks,
            linkcolor=red,
            anchorcolor=red,
            citecolor=green]{hyperref}
\usepackage{cite}
\usepackage[capitalize]{cleveref}
\crefname{section}{Sec.}{Secs.}
\Crefname{section}{Section}{Sections}
\makeatletter
\let\MYcaption\@makecaption
\makeatother

\usepackage[font=normalsize,labelfont=sf,textfont=sf]{subcaption}

\makeatletter
\let\@makecaption\MYcaption
\makeatother

\def\ourmodel{Fed-NCL}

\begin{document}

\title{Federated Noisy Client Learning}

\author{Kahou Tam, Li Li, Bo Han, Chengzhong Xu,~\IEEEmembership{Fellow,~IEEE}, and Huazhu Fu,~\IEEEmembership{Senior Member,~IEEE}
\thanks{K.~Tam, L.~Li, C.~Xu are with State Key Laboratory of Internet of Things for Smart City, University of Macau, Macau, China (email: wo133565@gmail.com; llili@um.edu.mo; czxu@um.edu.mo).}
\thanks{B.~Han is with Department of Computer Science, Hong Kong Baptist University, HongKong, China (email: bhanml@comp.hkbu.edu.hk).}
\thanks{H.~Fu is with Institute of High Performance Computing, Agency for Science, Technology and Research (A*STAR), Singapore (email: hzfu@ieee.org).} 
\thanks{\textit{Corresponding author: Huazhu Fu.}}
}

\markboth{SUBMITTED FOR REVIEW}%
{Shell \MakeLowercase{\textit{et al.}}: A Sample Article Using IEEEtran.cls for IEEE Journals}


\maketitle

\begin{abstract}
	Federated learning (FL) collaboratively trains a shared global model depending on multiple local clients, while keeping the training data decentralized in order to preserve data privacy. However, standard FL methods ignore the noisy client issue, which may harm the overall performance of the shared model.
	We first investigate critical issue caused by noisy clients in FL and quantify the negative impact of the noisy clients in terms of the representations learned by different layers.
	We have the following two key observations: (1) the noisy clients can severely impact the convergence and performance of the global model in FL, and (2) the noisy clients can induce greater bias in the deeper layers than the former layers of the global model. 
	Based on the above observations, we propose Fed-NCL, a framework that conducts robust federated learning with noisy clients. Specifically, Fed-NCL first
	identifies the noisy clients through well estimating the data quality and model divergence. Then robust layer-wise aggregation is proposed to adaptively aggregate the local models of each client to deal with the data heterogeneity caused by the noisy clients. We further perform the label correction on the noisy clients to improve the generalization of the global model.
	Experimental results on various datasets demonstrate that our algorithm boosts the performances of different state-of-the-art systems with noisy clients. 
	Our code is available on \url{https://github.com/TKH666/Fed-NCL}.
\end{abstract}

\begin{IEEEkeywords}
Federated Learning, Noisy Client, Label Noise, Noisy Learning.
\end{IEEEkeywords}

\section{Introduction}
\IEEEPARstart{L}{ocal} clients, such as personal devices, financial institutions, or hospitals, have access to a wealth of private data. However, due to data privacy concerns, security limitations, and device availability, it is impractical to collect and store all the data from the local clients at the server center and conduct centralized training. To ensure data privacy in the learning procedure, federated learning (FL) is recently proposed as a workflow that enables local clients to collaboratively train a shared global model without sharing local private data~\cite{Konecny2016,Yang2019,Li2020_SPM,Kairouz2019}. 
Figure 1 represents the workflow of a typical FL framework which can be mainly divided into the following two stages: 1) the client performs local training independently based on its local data, and the trained local model is sent back to a central server, 2) the central server then aggregates the local gradient updates and generates the global model. This process iterates until a certain accuracy level of the learning model is reached. In this way, an accurate predictive model can be obtained while the user data privacy is effectively protected, as the local training data are not accessed directly. 
Thus, FL has been widely adopted to support different application scenarios with high requirements of data security and privacy, such as smart retailers collecting user purchasing records~\cite{Tan2020,MobiCom20}, hospitals containing a multitude of patient data for predictive health care~\cite{Rieke2020,Bercea2022,Li2020} and financial institutions storing highly personal data for risk prediction~\cite{Byrd2020}.

\begin{figure}[!t] 
	\centering
	\includegraphics[width=1\linewidth]{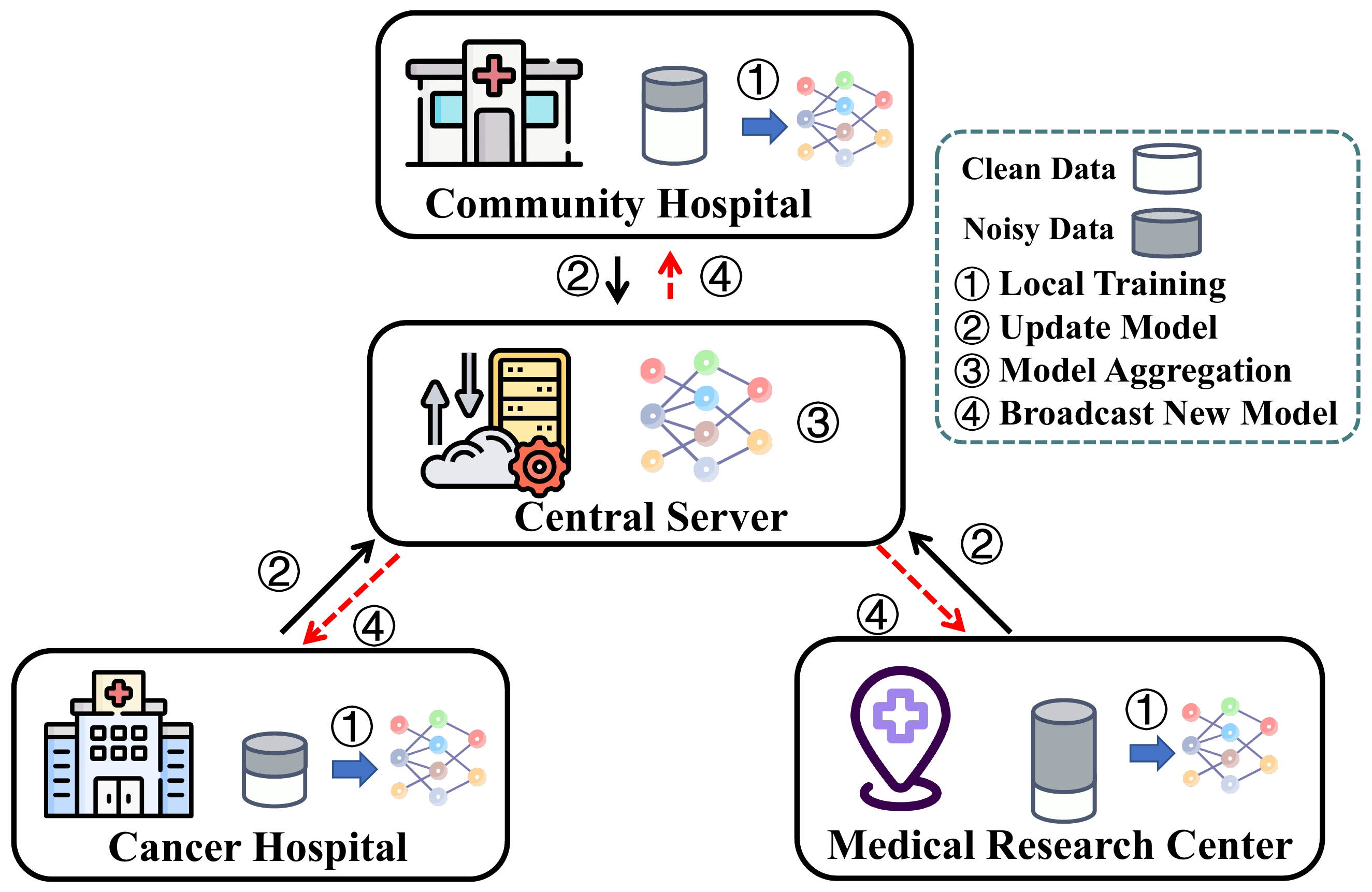}
	\caption{A typical FL architecture over server-client infrastructures, where the central server aggregates the local gradient updates from clients, keeping the local training data decentralized. It is difficult to guarantee that the data collected from different institutions are clean. Different from noisy label learning, federate noisy client learning focuses on the global noisy client distribution.}
	\label{fig_cover} 
\end{figure}

Despite all its promise, one critical obstacle exists for FL to be viable in real-world scenarios. In general, the performance of machine learning highly depends on the data quality of the participating clients~\cite{NIPS2013_3871bd64,Yu2019,Zhang2019b}. However, carefully labeled data are usually expensive and time-consuming to collect. The data annotation process is also extremely complex even for experienced people. It is difficult to guarantee that the labels of the training data located on each client are clean, as it is impossible to ensure that all the clients can perfectly annotate the corresponding training data. Training with noisy data (\textit{e.g.}, feature noise and label noise) reduces the generalization performance of the deep learning model, since the networks can easily overfit on corrupted data~\cite{pmlr-v70-arpit17a,Zhang2017}. 
The noisy data issue is even more serious in the FL setting, which collaboratively aggregates the global models depending on a large number of clients during the overall training process. The clients usually have entirely different standards for collecting their local data, and may contain some non-expert and uncertain sources. The use of these noisy clients often results in unreliable labels and harms the global aggregated model.
Unfortunately,  standard FL aggregation methods, \textit{e.g.}, FedAvg~\cite{FedAvg_McMahan2017} and FedProx~\cite{FedProx_Li2018}, ignore the noisy client issue and develop the global model directly based on the trained model from each participating client.
Thus, the overall performance of FL can be negatively impacted by the noisy clients. 

Recently, many machine learning (ML) techniques have been developed to handle noisy data~\cite{6685834,Song2020} in the centralized setting.  The state-of-the-art noisy learning methods have shown significant improvement by re-weighting training samples~\cite{pmlr-v37-menon15,7159100} or applying a disagreement constraint between different networks/training strategies~\cite{Decoupling,MentorNet,CoTeaching}. However, these noisy learning methods cannot be applied in FL workflow directly. 
First, the local training data is inaccessible to the central server. Thus, it cannot remove/re-weight the noisy data during the global model aggregation.
Second, different from the single dataset in standard ML, FL has several independent local clients, which have different noisy data distributions. This makes it difficult to estimate the unified noise transition matrix used in noisy learning methods~\cite{pmlr-v37-menon15,7159100}. 
Third, some clients (\textit{e.g.}, battery-powered mobile devices) in FL may have limited computing resources, so that the specific noise adaptation layer~\cite{adaptation-layer} or additional collaborative network~\cite{CoTeaching} cannot be satisfied. 

Training the collaborative model with noisy label data in FL has recently gained more attention. Some methods~\cite{chen2020focus,tuor2021overcoming} utilize the extra clean data to identify the clients with noisy labels and mitigate their negative impact by re-weighting during the global model aggregation. Yang \textit{et al.}~first select the confident data to train the local model and perform label correction on noisy label data~\cite{yang2022robust}. However, these existing methods are still not effective and stable in the real-world setting~\cite{li2021federated} for the following reasons. 
First, from the perspective of noisy clients, they assume all the clients have the same noise degree instead of heterogeneous noise distribution, which is not practical in the real-world scenario. 
Second, they rely on the benchmark data to detect the clients with noisy label data and select the confident training data of noisy clients, which is hard to satisfy in the real-world scenario since it restricts the task-generalization of FL. 
Third, most methods only consider the effectiveness without understanding how the noisy clients affect each layer of the global model. 
Thus, a framework that can effectively identify noisy clients and conduct robust learning without relying on benchmark data is urgently required for FL. 

In this paper, we first analyze the problem of noisy client in FL\footnote{In generally, there are two types of noise as: feature and label noises~\cite{6685834}. Here, we only consider the label noise case, because the feature noise could be treated with the similar way.}, model the noise levels among the clients with two distributions (e.g., Bernoulli and truncated Gaussian distributions) to better emulate heterogeneous noisy scenarios in the real-world federated system. Then, to identify how the noisy clients affect the global model in FL,  we first perform the experimental study on the performance of the global model and convergence rate when the noisy clients exist. We observe that the noisy data existing in participating clients can negatively impact the convergence and performance of the aggregated global model in FL. Moreover, we dig deeper to investigate how noisy clients affect each layer of the global model in federated learning. Specifically, we utilize the Centered Kernel Alignment (CKA)~\cite{kornblith2019similarity} to measure each layer's output feature similarity between the noisy clients and the global model. The observation is that the noisy clients can induce greater bias in the deeper layers than the former layers of the global model when directly using the FedAvg~\cite{FedAvg_McMahan2017} to aggregate all clients' models.

Based on the above observation, we propose Fed-NCL, a Federated Noisy Client Learning framework, which effectively tackle the noisy clients in FL without relying on the benchmark data.
In \ourmodel, we detect noisy clients through modeling the reliability score distribution among the participating clients. The reliability score jointly considers each client's local model divergence with global model and its data quality. Based on the observation that the deeper layer's feature extraction ability in the noisy clients' model heavily declined, a noise robust layer-wise aggregation is designed to aggregate the feedback model from each client by considering each layer's divergence and guiding the global model update's direction like clean clients. In order to fully utilize the local data from noisy clients to improve the generalization of the global model, we further perform the label correction on the noisy clients. 
In summary, the main contributions of this paper are as follows: 
\begin{itemize}
	\setlength{\itemsep}{5pt} 
	\setlength{\parskip}{0pt}
	\item We present the first systematic and empirical study on federated learning with noisy label data and analyze the impact of the hidden representations of different layers of deep neural networks (DNN) trained with FedAvg on noisy label data.
	\item We propose two types of noisy label scenarios in the real-world federated system by modeling the noise level among the clients with two distributions (\textit{e.g.}, Bernoulli and truncated Gaussian distributions).
	\item Our systematic study on federated learning with noisy clients reveals that (1) the noisy clients could negatively impact the convergence and performance of the global model in FL, and (2) the noisy clients can induce greater bias in the deeper layers than the former layers of the global model.  
	\item We propose \ourmodel, which dynamically identifies the noisy clients by reliability score without making any assumptions on the noisy clients. The reliability score measures each client's data quality and model divergence with the global model. Then
	we develop a robust layer-wise aggregation methodology, which mitigates the negative impact from the noisy clients in term of layers. The label correction is dynamically performed in the noisy clients to improve the generalization capabilities of the global model.
	\item Experimental results on various datasets demonstrate that our algorithm can boost the robustness and performances of different deep learning backbones in FL systems against noisy clients.  
\end{itemize}

\section{Related Works} 
\label{sec_related}

\subsection{Deep Learning with Noisy Label}

Several methods have been designed to mitigate the impact of noisy data in deep learning~\cite{6685834, Song2020,pmlr-v37-menon15,7159100, Decoupling,MentorNet,CoTeaching, adaptation-layer}. 
Goldberger \textit{et al.}~\cite{adaptation-layer} proposed to model the data noise with an additional softmax layer during the training process. 
Liu \textit{et al.}~\cite{pmlr-v37-menon15} re-weighted the importance of the training samples in order to reduce the impact of noisy labels. 
Jiang \textit{et al.}~\cite{MentorNet} designed a collaborative learning framework (MentorNet) to introduce a pre-trained mentor network that directs the training process. 
In addition, Han \textit{et al.}~\cite{CoTeaching} also proposed to maintain two networks, where each network selects the training samples that have a small loss and passes them as input to its peer network to continue the training process. 
However, these existing approaches cannot be directly adopted in FL scenarios for the following two reasons. First, in order to preserve data privacy, the central server cannot access the raw training data located on each client. Second, the computational overhead caused by introducing an extra network in the training process can highly impact the energy efficiency and user experience for battery-powered clients in FL (\textit{e.g.}, smartphones and wearable devices).

\subsection{Federated Learning with Noisy Label}
Existing works about FL with noisy labels can be summarized into the following two main categories: 1) Benchmark data-dependent methods and 2) Label correction-based methods. The Benchmark data-dependent methods extract the subset of clean clients or data with clean labels using public benchmark data. 
FOCUS \cite{chen2020focus} exploits the benchmark data stored in the server to evaluate all participant clients' models to adjust their weight of global model aggregation.
DS \cite{tuor2021overcoming} selects the confident samples by estimating the similarity between the client's training data and benchmark data from the server. The label correction-based methods first train the global model with the client's confident samples and then performs the label correction on the client's data with noisy labels. For instance, RoFL \cite{yang2022robust} utilizes label correction while training the sample with small loss to create local centroids and exchange them between clients   and servers. CLC \cite{xu2022fedcorr} utilizes the consensus-defined class-wise information to identify the noisy labels data on the clients and correct them. 

However, these methods have the following critical limitations. First, they don't consider the heterogeneous noise distributions across different clients, assuming all the clients have the same noise degree.
Second, some approaches \cite{tuor2021overcoming,chen2020focus} have the strong assumptions that the server exists the benchmark data, which is hard to satisfy in real-world applications.
Moreover, the above methods only extend the existing methods \cite{CoTeaching, MentorNet, Decoupling} from centralized learning to the federated setting, they lack a deep understanding of how the noisy clients affect each layer of the global model in FL. Our work can help to discover and resolve the unique challenge of noisy clients in FL.

\section{Problem Statement}
\label{sec_noisy}

Several label noise distributions have been effectively modeled for standard ML~\cite{6685834,Ortego2019,NEURIPS2018_aee92f16}, \textit{e.g.}, uniform or non-uniform random label noise. However, the noise scenario in FL is different from the label noise in standard ML. The noise modeling of standard ML only focuses on a single training dataset, while FL has multiple clients with their own private datasets. Thus, we focus on the noise distribution among all the participating clients.
In this section, we present the problem statement of federated learning with noisy clients, and model the noisy clients with different distributions.

\subsection{Federated Learning with Noisy Client}
\label{sec_FLinNC}

In this paper, we consider the $K$-class image classification, which is a representative supervised learning task and 
can be formulated to learn the mapping function $f(x; \Theta)$ from a set of training examples $D=\{(x_{i}, y_{i})\}^{N}_{i=1}$ with $y_i \in \{0,1\}^{K}$  being the ground-truth label in a one-shot manner corresponding to $x_i$. In deep learning, $f(x; \Theta)$ is a network and $\Theta$ represents the model parameters.  
The parameter $\Theta$ minimizes the empirical risk $\mathcal{R}(f)$, as: 
\begin{equation}
\mathcal{R}(f)=\sum_{(x_i, y_i) \in D}^{N}l(f(x_i;\Theta),y_i),
\end{equation} 
where $l(\cdot)$ is a loss function (\textit{e.g.}, cross-entropy loss or mean squared error). 

In the FL workflow, as data labels are corrupted in the training data of the participating clients, we aim to train the model from noisy clients. Specifically, a certain client $c$ is provided with a noisy training dataset $\tilde{D}^c=\{(x_{i}^c, \tilde{y}_{i}^c)\}^{N}_{i=1}$,
where $\tilde{y}_{i}^c$ is a noisy label. Hence, following the standard training procedure, a mini-batch $ \mathcal{B}^c_{t}=\{(x_{i}^c,\tilde{y}_i^c)\}^{b}_{i=1}$ comprising $b$ samples is obtained randomly from the noisy training dataset $\tilde{D}^c$ of client $c$ at time $t$. Subsequently, the local model parameter $\Theta_{t}^c$ of client $c$ at time $t$ is updated along the descent direction of the empirical risk on mini-batch $\mathcal{B}^c_{t}$: 
\begin{equation}
\Theta_{t+1}^c = \Theta_{t}^c-\eta^c \nabla \left[  \sum_{(x_{i}^c, \tilde{y}_{i}^c) \in \mathcal{B}^c_{t} } l(f^c(x_i^c;\Theta_{t}^c), \tilde{y}_i^c) \right],
\end{equation} 
where $\eta^c$ is a specified learning rate in client $c$. Thus, the risk minimization process is no longer noise-tolerant because of the loss computed by the noisy labels in the local training process for the noisy clients. 

In a training round, participating clients send their updated model parameters to the central server after completing the local training. The central server then aggregates the global model $\Theta^G$ by federated averaging~\cite{FedAvg_McMahan2017}, as: 
\begin{equation}
\Theta^G = \dfrac{1}{C} \sum_{c=1}^{C} \Theta^{c},
\label{eq_fl_aggre}
\end{equation}
where $C$ represents the total amount of local clients in the current training round, and $\Theta^{c}$ represents the model parameter of client $c$ in the current training round. 
However, some of the participating clients are noisy clients, which means that some of the local models have memorized corrupted labels. Thus, directly aggregating these models using Eq.~(\ref{eq_fl_aggre}) leads the update of the global model to a divergent direction, and the generalization of the global model is severely impacted correspondingly. Hence, mitigating the negative effects of noisy clients is critical to make federated learning viable in practice.

\subsection{Modeling Noisy Client}

In this section, we discuss how to model the noisy client scenario with two distributions, \textit{i.e.}, Bernoulli and truncated Gaussian distributions.

\noindent\textbf{Case 1: Bernoulli Distribution.} In this case, we use Bernoulli distribution to model the noisy clients across the participants in a federation. The Bernoulli distribution is a discrete distribution having two possible outcomes labeled by $n=0$ and $n=1$, in which $n=1$ (the training data of a specific client are clean) occurs with probability $p$ and $n=0$ (the training data of a specific client are noisy) occurs with probability $q=1-p$, where $0<p<1$. The probability density function $P_B$ is represented as:
\begin{equation}
P_B(n) = p^{n}(1-p)^{1-n}.
\end{equation}
It is important to note that, in this case, we assume that if a client is a noisy client ($n=0$), the labels of all the training data within it are corrupted (\textit{e.g.}, some sensors on certain mobile clients malfunction and generate corrupted training data). Otherwise, if a client is a clean client ($n=1$), the labels of all the training data have true labels. For the reason that whether the data collection component in client (\textit{e.g.}, sensors on mobile devices) malfunctions or not is entirely independent from each other and follows the Bernoulli process, thus Bernoulli distribution~\cite{bernoulli} is adopted in this case.


\noindent\textbf{Case 2: Truncated Gaussian Distribution.} In this case, the local training data of all the clients in federated learning may have labels corrupted to different degrees (\textit{e.g.}, different hospitals may collect data with various qualities). We use a truncated Gaussian distribution~\cite{trungaussian} to model the percentage of corrupted labels within the training data across different clients. Specifically, the truncated normal distribution is derived from a normally distributed random variable by giving the random variable a lower and upper bound. Specifically, suppose $X$ has a normal distribution with mean $\mu$ and variance $\sigma^{2}$ and lies between $(a,b)$. Then $X$ follows a truncated normal distribution conditioned on $a < X < b$. The probability density function $P_G$, for $a\le x \le b$ can be represented as follows: 
\begin{equation}
P_G(x;\mu, \sigma, a, b) = \frac{1}{\sigma}\frac{\phi(\frac{x-\mu}{\sigma})}{\Phi(\frac{b-\mu}{\sigma})-\Phi(\frac{a-\mu}{\sigma})},
\end{equation}
where $\phi (\cdot)$ is the probability density function of the standard normal distribution, and $\Phi (\cdot)$ is its cumulative distribution function. $x$ is defined as the noise degree for a certain client, which can be represented as $x=\frac{n_{noise}}{n_{clean} + n_{noise}}$. $n_{noise}$ is the amount of noisy data, while $n_{clean}$ represents the amount of clean data within a certain client. Moreover, $x$ is bounded between 0 and 1.


\begin{figure}[!t]  
	\centering 
    \includegraphics[width=1\linewidth]{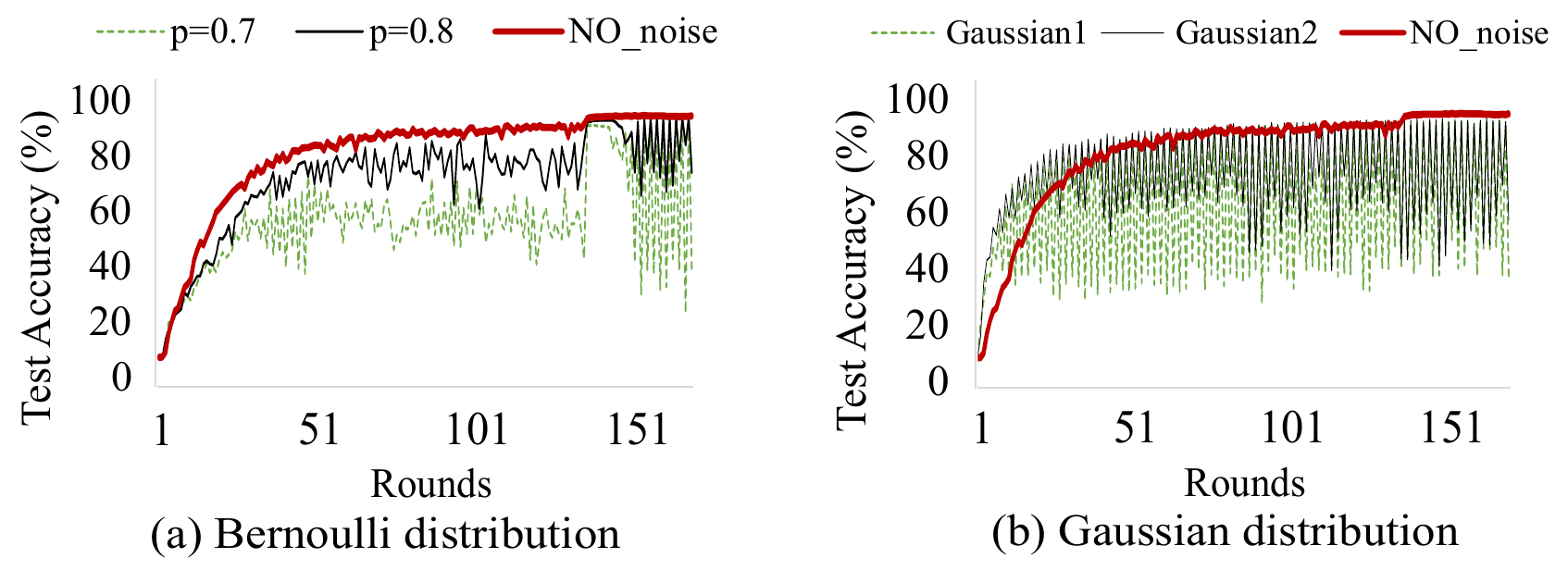}
	\caption{The performances of VGG16 on CIFAR10 with different noisy client distribution.(a) Bernoulli distribution with $p=\{ 0.7, 0.8\}$. (b) Gaussian Distribution (Gaussian 1: $\mu=0.3, \sigma = 0.45$; Gaussian 2: $\mu=0.3, \sigma = 0.4$).}
	\label{resnetrealexample} 
\end{figure}

\noindent\textbf{A Real-World Example.}
We now use a real-world example (training VGG16~\cite{VGG} on CIFAR10~\cite{cifar10}) to show how the noisy clients impact the performance of federated learning. Because the original dataset is clean, we manually corrupt it with an approach named symmetric noise~\cite{Song2020}. Specifically, the true label is corrupted by a label transition matrix $T$, where $T_{ij}:=p(\tilde{y}=j|y=i)$ is the probability of the true label $i$ being flipped into a corrupted label $j$. The true labels are flipped into other labels with equal probability.
Fig.~\ref{resnetrealexample} shows the test accuracy versus the communication rounds for VGG16 on CIFAR10 in the above two cases: 

\noindent{\textit{1) Bernoulli distribution:}} For noisy clients following the  Bernoulli distribution, we test two different scenarios: $p=0.7$ and $p=0.8$. Fig.~\ref{resnetrealexample} (a) shows the test accuracy with different communication rounds. As can be seen, the training process can smoothly converge in a few communication rounds (\textit{e.g.}, before round 150 in this case). However, the convergence process slows down prominently when $p=0.8$ ($20\%$ of the clients are noisy clients). In addition, the situation gets even worse when the percentage of noisy clients increases (\textit{e.g.}, $p=0.7$). This is because the local models located at the noisy client try to memorize the corrupted data in the local training process. In addition, during the model aggregation process, these local models will guide the update of the collaborative model in a divergent direction. 

%


\begin{figure}[!t] 
	\centering
	\includegraphics[width=1\linewidth]{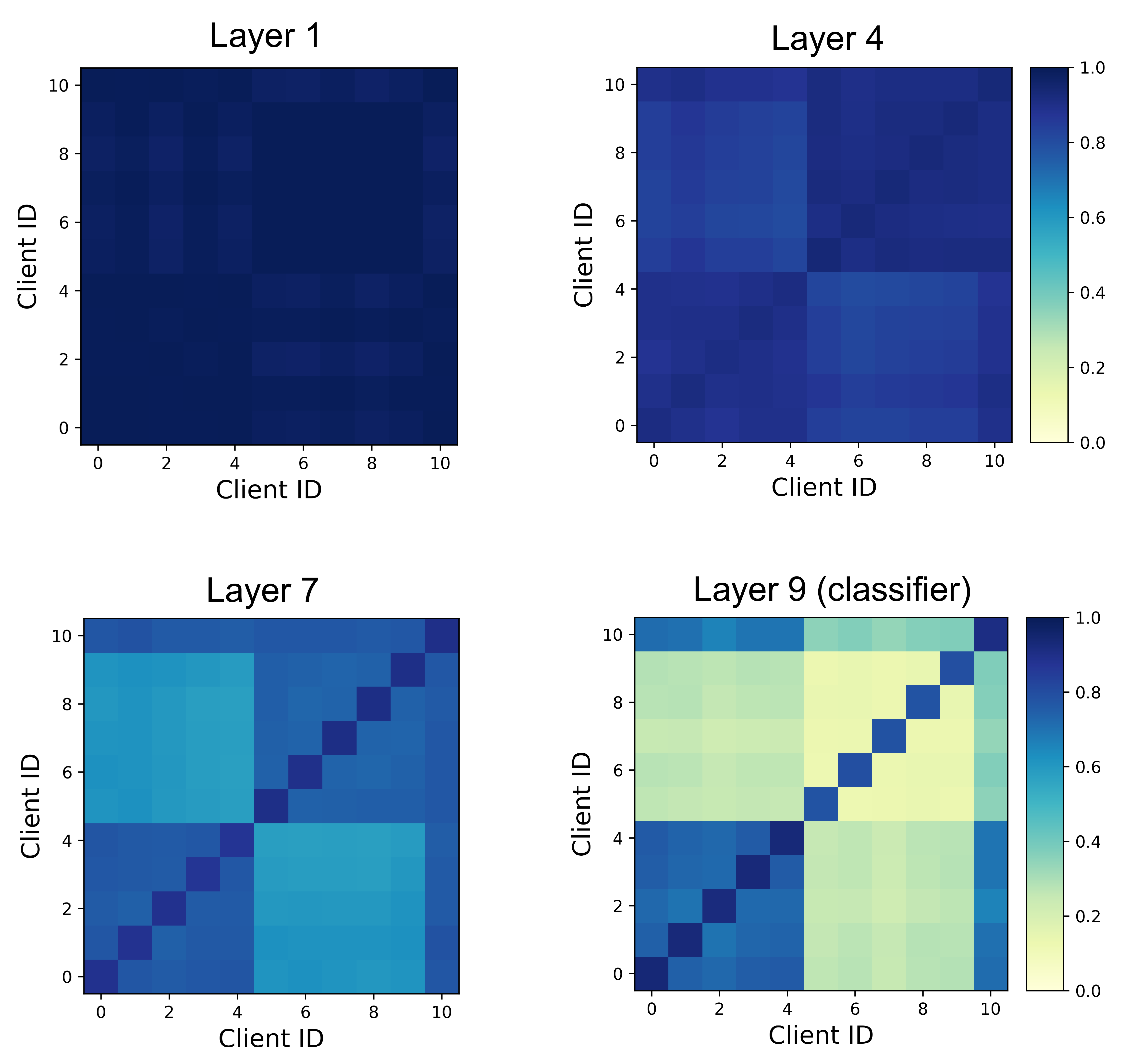}
	\caption{The CKA similarities of three different layers among noisy and clean clients. Clients 0-4 are clients with clean data. Clients 5-9 are noisy clients. Client 10 is the global model.}
	\label{cka} 
\end{figure}

\noindent{\textit{2) Truncated Gaussian distribution:}} 
For noisy clients following the truncated Gaussian distribution, we test the following two scenarios: ($\mu=0.3, \sigma = 0.45$) and ($\mu=0.3, \sigma = 0.4$). Fig.~\ref{resnetrealexample} (b) represents the test accuracy during the overall training process. As can be seen, the truncated Gaussian distributed noisy clients also negatively impact the federated learning process. Moreover, we can notice that the impact increases as the standard deviation of the distribution increases. 

\section{A Close Look at Noisy Client}\label{sec_closelook}
To investigate how the noisy clients affect the classiﬁcation model in standard federated learning procedure \cite{FedAvg_McMahan2017}, we conduct the following experiment. We use 10 clients to train the 9-layer convolutional neural network used in \cite{han2018co} on the CIFAR-10. We follow the i.i.d setting to partition the data into 10 clients. In addition, we set 5 clients as noisy clients whose training data are manually injected with the 50\% symmetric label noise. The rest of the clients have data with clean labels. To specifically show how the noisy clients impact each layer of the global model, we measure each layer's representations extraction ability by computing the Centered Kernel Alignment (CKA) \cite{kornblith2019similarity} similarity between the representations from the same layer of different clients' local models. 
The CKA is a representation similarity metric that is widely used to compute the feature representation learned by two neural networks \cite{nguyen2020wide,luo2021no}.
In our experiments, we use linear CKA to calculate the similarity of the output features between the two models. Given the same dataset $D_{cka}$, feature matrix $Z_1, Z_2$ can be extracted by two models, respectively. Then we can calculate the linear CKA similarity between two representations $X$ and $Y$ as:
\begin{equation}
CKA(X,Y)=\frac{||X^TY||^2_F}{||X^TX||^2_F||Y^TY||^2_F} ,
\end{equation}
The CKA similarity score is between 0 (not similar) and 1 (identical) to show the output feature similarity of the same layer across two local models. We train the global model using the FedAvg \cite{FedAvg_McMahan2017} for 100 communication rounds, and the clients locally train for 10 epochs at each round. 

\begin{figure}[!t] 
	\centering
	\includegraphics[width=1\linewidth]{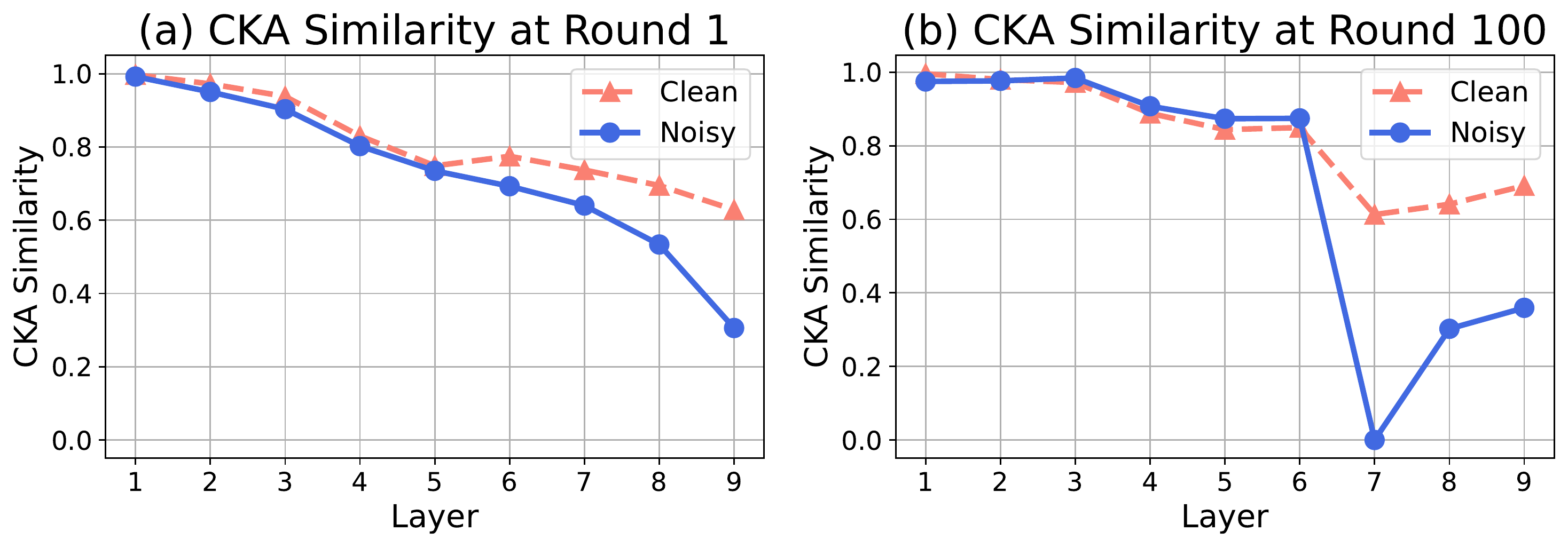}
	\caption{The means of the CKA similarities of different layers between the global models and the noisy/clean clients.}
	\label{mean_cka} 
\end{figure}

Fig.~\ref{cka} shows the pairwise CKA features similarity score of the four different layers among the clients' local models and the global model. Four layers including the model's 1st, 4th, 7th, and last layer (classifier layer) are selected to show. Interestingly, we find that the deeper layer in the model has a lower CKA feature similarity score between noisy and clean clients. This indicates that the deeper layers represent higher heterogeneity across the noisy and clean clients in federated noisy client learning. 
By averaging each layer's CKA similarity between the global model and clean/noisy clients' local models respectively, we obtain the approximate of each layer's feature output similarity between the global model and noisy/clean clients' models.
Fig.~\ref{mean_cka} represents the experiment result which shows that the noisy clients' models have consistently lower CKA similarity score as the layer becomes deeper, compared with the global model. This lower CKA feature similarity score between the global model and noisy clients indicates that the local model from the noisy clients diverges from the global model in terms of the feature extraction ability. Although noisy clients exist, the global model's feature extraction ability still has high consistency with clients with clean labels, reflecting a high CKA similarity score with clean clients. 
Based on the above experiment study on federated noisy client learning, we can observe that compared with training on the clients with clean labels, the deeper layer's feature extraction ability in the noisy client's model is severely declined, reflected by the lower feature similarity among the noisy and clean clients' deeper layers. Moreover, the deeper layer of the noisy clients' models can be biased by their noisy label data, causing divergence from the global model.

In summary, the current federated learning workflow is not robust to noisy clients. The model convergence, convergence rate, accuracy, and feature extraction ability can all be negatively impacted by noisy clients. However, noisy clients commonly exist in different real-world application scenarios. In order to make federated learning viable in practice, a robust framework for noisy clients is urgently required.

\begin{figure*}[!t] 
	\centering
	\includegraphics[width=1\linewidth]{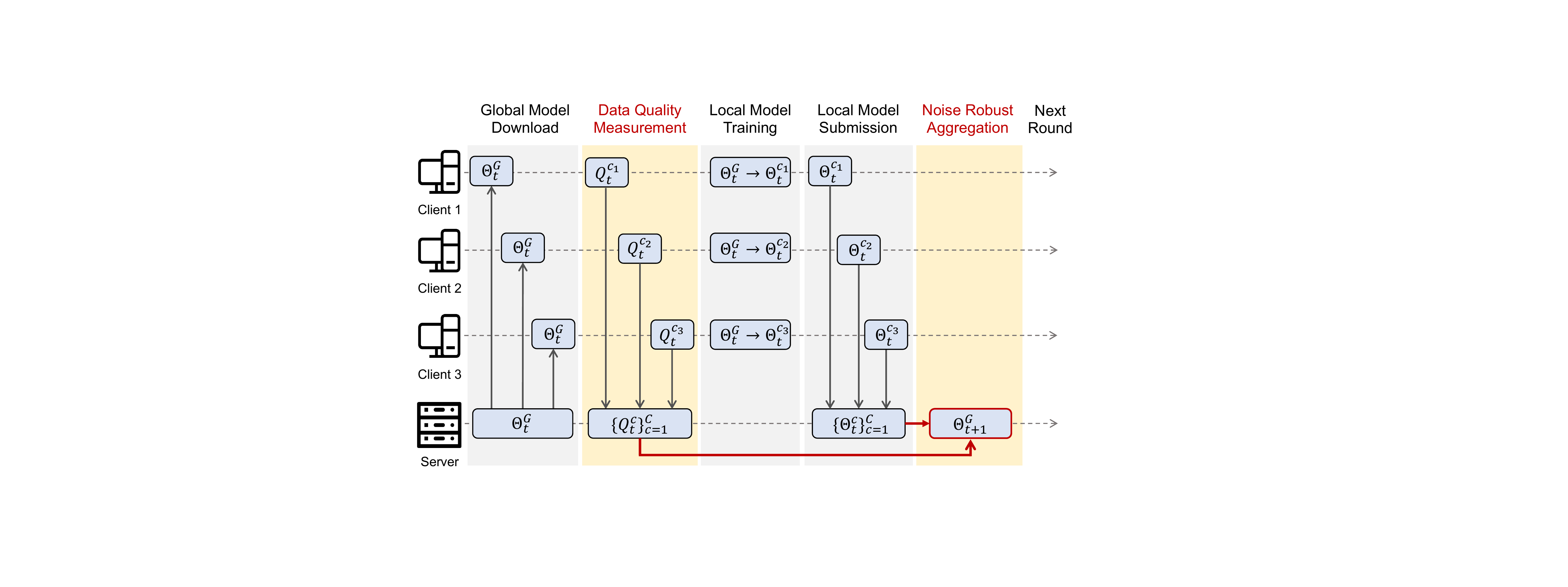}
	\caption{The workflow of our \ourmodel. The workflow contains 5 steps: 1). the server sends the global model $\Theta_G$ to each client;  
	2). local training on private data;
	3). each client sends its local model and loss to the server;
	4). the server detects noisy clients by measuring the clients' reliability scores.
	5). the server performs layer-wise aggregation by weight matrix $W_t$.
	}
	\label{workflow} 
\end{figure*}

\section{Federated Noisy Client Learning} 
According to the observation in Sec.~\ref{sec_closelook}, directly aggregating the local model trained from noisy clients can hurt the performance of the global model. Hence, it is critical to identify noisy clients and mitigate their negative effects. There are two main technical challenges for FL with noisy clients: 1) how to identify the client with noisy data while preserving data privacy, and 2) how to conduct a robust model aggregation to effectively mitigate the impact of noisy clients. In this section, we present the design of Federated Noisy Client Learning (Fed-NCL), which effectively distinguishes noisy clients and intelligently mitigates the impact of noisy clients during the overall FL process. \cref{workflow} is an overview of Fed-NCL which mainly contains the following three stages:  1) identification of noisy clients, 2) robust layer-wise adaptation aggregation, and 3) label correction. In identifying noisy clients, the server calculates the reliability scores of the clients to find out the noisy clients statistically. After detecting the noisy clients, the server performs robust layer-wise adaptation aggregation, which jointly considers the model's layer divergence and the impact of noisy clients, to obtain a global model for the next round of local training. Finally, we correct the label from the noisy clients to reduce the negative impact of noisy clients and gain more valuable features from the noisy clients. 

\subsection{Noisy Client Detection.\label{noisydetection}}
Identifying noisy clients is crucial to preventing the noisy client model from damaging the global model aggregation. Existing methods~\cite{tuor2021overcoming,chen2020focus} depend on the benchmark data, which is a set of data with completely clean labels stored on the server. The requirement of such methods is difficult to satisfy in real-world applications because the clean label data is hard to retrieve in certain application scenarios. Therefore, we propose a benchmark data-free noisy client detection approach. The key idea of our noisy client detection approach is that each participant is assigned a reliability score which is quantified based on the client's data quality and the divergence between the client and global models. Then, the server utilizes the abnormal detection model to identify the noisy clients based on the distribution of all clients' reliability scores.

\begin{figure}[!t] 
	\centering
	\includegraphics[width=1\linewidth]{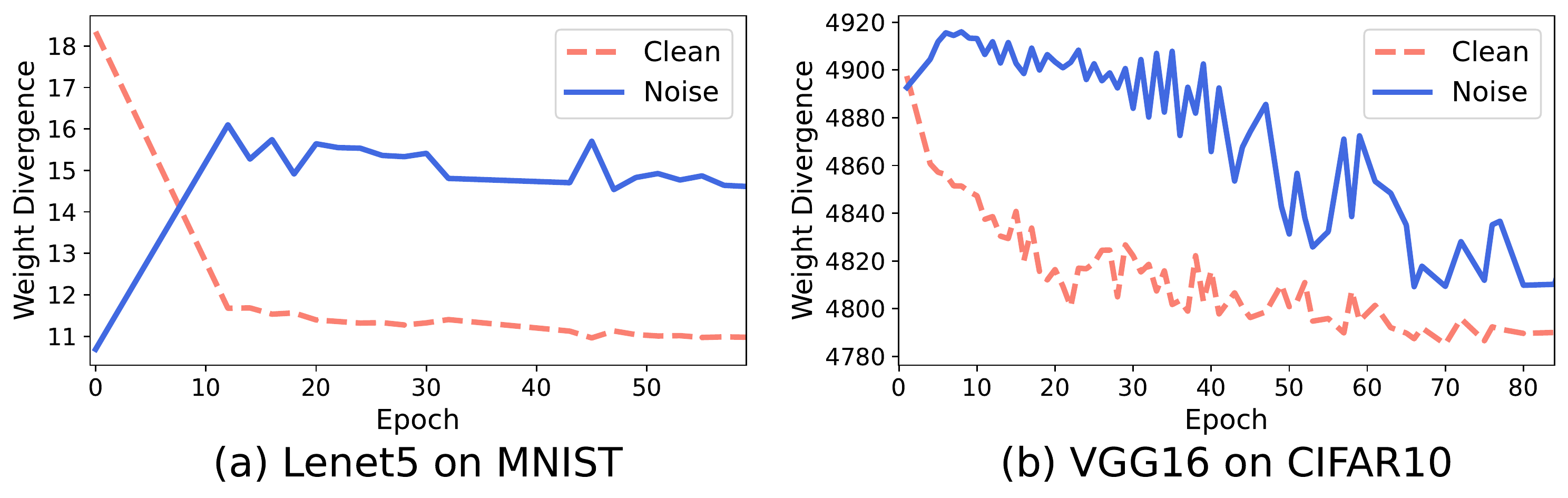}
	\caption{The weight divergence between the noisy clients and global model during training.}
	\label{weightdivergence} 
\end{figure}

\textbf{Reliability Score.} 
To effectively detect noisy clients, we measure each client's model with the reliability score, which jointly considers its data quality and model divergence with the global model. 

According to the experiment study between the noisy clients and clean clients in Sec.~\ref{sec_closelook}, we observe that the model trained on the noisy clients represents heavier heterogeneity in feature extraction from the global model. This heterogeneity indicates that the noisy clients' local model significantly diverges from the global model. Based on this observation, we measure the model divergence $e$ between the global model and client $c$ at training round $t$ as:
\begin{equation}
e_t^{c} = ||\Theta^{G}_{t}-\Theta^c_{t}||^2 ,
\label{modeldistance}
\end{equation}
where $\Theta^{G}_{t}$ represents the set of parameters of the aggregated global model with model averaging~\cite{FedAvg_McMahan2017} in the current training round t, while $\Theta^c_{t}$ is the set of parameters of the local model on client $c$. \cref{weightdivergence} shows the divergence between the global and local models across the clean and noisy clients.
As the training proceeds, the local model of a clean client and the updated global model has a small distance. However, the distance between the noisy client's local model and the clean client's local model keeps at a high level. Thus, model distance is an effective metric to quantify the model quality in FL. 

To quality each client's data quality, we use the cross-entropy (CE) loss of the local training data on the collaborative model:
\begin{equation}
h_t^{c} =\sum_{(x_i, y_i) \in D^c}^{N}\text{CE}(f(x_i;\Theta_t^{c}),y_i),
\label{CEloss}
\end{equation}
In deep learning, it has been empirically shown that deep networks can learn the clean pattern at the start of training \cite{han2018co,arpit2017closer,chen2019understanding,zhang2021understanding}. Thus, the deep networks can discriminate the noisy and clean label data, reflecting the prediction with high confidence on clean label data. Therefore, the clients with clean label data should have low empirical risk.

\begin{figure}[!t] 
	\centering
	\includegraphics[width=1\linewidth]{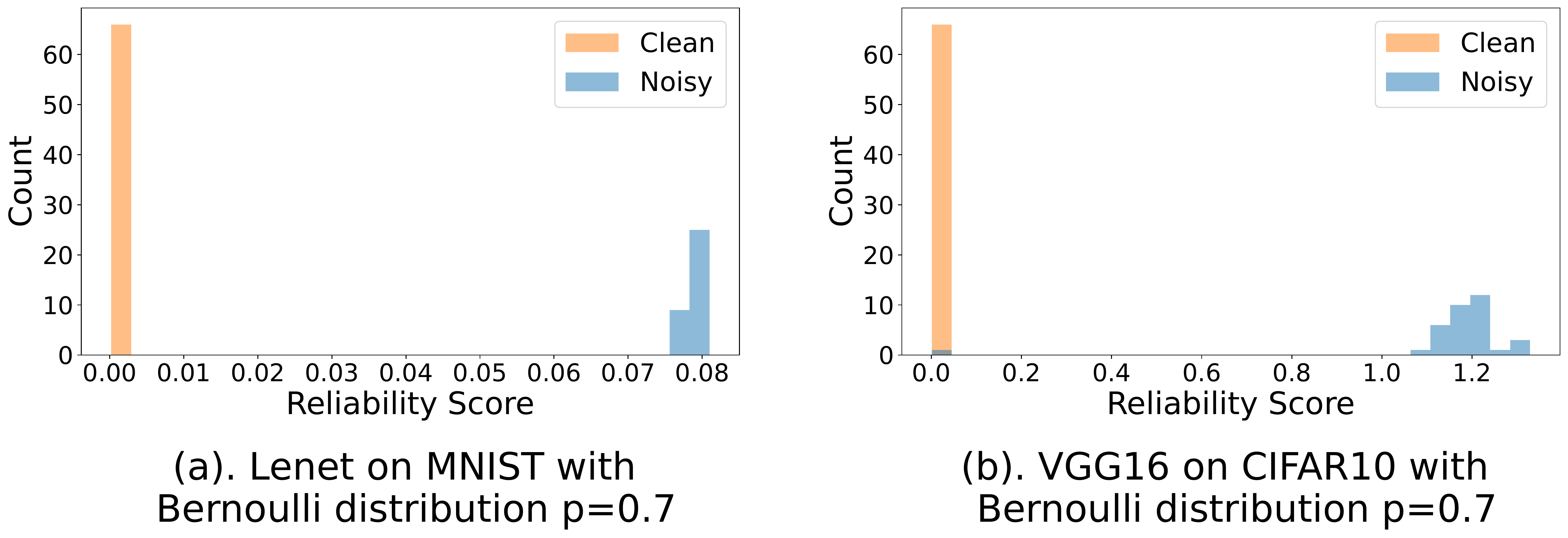}
	\caption{The distribution of reliability scores of all participant clients after 10 global rounds with Bernoulli noise model $p=0.7$, and with IID data partition, over 100 clients. (a): Training the Lenet on the MNIST. (b): Training the VGG16 on the CIFAR10. }    
	\label{clientreliability_score} 
\end{figure}

Finally, we integrate the two selection criteria in Eqs.~(\ref{modeldistance}) and (\ref{CEloss}) into our reliability score:
\begin{equation}
q_t^{c} = \frac{e_t^{c}* h_t^{c}}{|D^c|},
\end{equation} 
where $|D^c|$ is the amount of training data. The reliability score consists of two metrics to quantify the client's model, which does not rely on extra benchmark data. It fully utilizes the information of the local model while preserving the data privacy of the clients. \cref{clientreliability_score} shows the distribution of reliability scores across clean and noisy clients in different experimental scenarios. We can observe that there is a large margin between the two groups of clients. Hence, the reliability score is an efficient selection criterion to distinguish noisy client from clean clients.

\textbf{Noisy Clients Detection.}
After getting the reliability score of each client, the server can distinguish the noisy clients with reliability scores $Q_t=\{q_t^{1},...q_t^{c}\}$. We treat the identification of noisy clients from all clients' reliability scores as an abnormal detection task. It is intuitive to set a fixed threshold or train an anomaly detection model (\textit{e.g.}, autoencoder) to filter out noisy clients. However, since the training process of federated learning is highly dynamic and unpredictable, these methods are difficult to guarantee the performance of anomaly detection. Therefore, we utilize anomaly detection with the normal distribution to identify noisy clients. Specifically, we approximate all clients' reliability scores as normally distributed. In statistics, the normal distribution has the conventional heuristic rule that nearly all values (99.7\%) are taken to lie within three standard deviations $\sigma$ of the mean. Based on this rule, if client $c$ breaks this rule, it will be identified as noisy client:
\begin{equation}
    q_t^{c}- E(Q_t) > \beta \sigma(Q_t),
\end{equation}
where $\beta$ is the number of standard deviations of the mean. We can adjust the $\alpha$ to guarantee anomaly detection accuracy to adapt to the high dynamics of federated learning training. After applying this procedure, the set of clients $S$ is divided into two subsets: $S_n$ (noisy clients) and $S_c$ (clean clients).


\subsection{Noise Robust Layer-wise Aggregation.}
To mitigate the negative impact of noisy clients, we propose noise-robust layer-wise aggregation. Our noise-robust layer-wise aggregation applies an aggregation weight matrix $W_t$ at the server to reduce the negative effect from noisy clients in term of layer:
\begin{equation}
W_t=\begin{bmatrix}
\mathbf{w}_{1}^{t}, & \mathbf{w}_{2}^{t},& \cdots ,& \mathbf{w}_{l}^{t} 
\end{bmatrix}
=\begin{bmatrix}
w_{1}^{1} & w_{1}^{2}& \cdots & w_{1}^{C} \vspace{2pt}\\ 
w_{2}^{1} & w_{2}^{2} & \cdots & w_{2}^{C} \\
\vdots & \vdots & \ddots & \vdots \\
w_{L}^{1} & w_{L}^{2} & \cdots & w_{L}^{C}
\end{bmatrix},
\end{equation}
where $\mathbf{w}_{l}^{t}$ represents the aggregation weight vector of $l$-th layer at round $t$, while $w_l^c$ represents the aggregation weight for client $c$ in $l$-th layer. For each layer $l$ , $\sum_{c=1}^{C}w_l^c=1$.

As mentioned above, the feature extraction ability of deeper layers in the noisy client's model is heavily declined. This finding also points out that the negative impact of noisy clients on each layer of the model is different. Therefore, for each client $c$, we quantify the distance $d$ with the global model of each layer at round $t$ as follows:
\begin{equation}
    d_{l_t}^{c}=1+ ||\Theta_{l_t}^{G} -\Theta_{l_t}^{c} ||^{2}  ,  l\in \{1,...,L\},
\end{equation}
where $L$ is the number of layers, $l$ is the layer index of the model. According to the training data distribution on different clients, for each layer $l$, the distance $D^c$ to the global model at round $t$ is defined as: 
\begin{equation}
D_{l_t}^{c} = \frac{\frac{N^c}{d_{l_t}^{c}}}{\sum_{c=1}^{C} \frac{N^c}{d_{l_t}^{c}}} .
\end{equation}
It is intuitive to abandon the noisy clients' models during global aggregation because they destroy the optimization of the global model. However, this kind of method reduces the convergence rate of the global model since the global model only learns the data from clients with clean label data, leading to a low generalization of the global model. According to the above discussion,  the noisy and clean clients shows that the output feature of the former layer of the local models from noisy and clean clients has high similarity. This implies that the noisy clients' local models can still learn the same pattern as the clean clients, and the noisy clients' models can contribute to the global model. Therefore, we use the $\tau$ to increase the distance between the global model and noisy clients' models while reducing the negative impact from noisy clients and improving the generalization of the global model. Even with noisy clients, they still can learn clean and easy patterns at the beginning of the training due to the memorization effect of the deep neural model. The problem is that the noisy clients' local model will gradually overfit their local noisy label data and destroy the global model. To address this problem, we design an adaptive $m(c)$ to control the penalty for noisy clients:
\begin{equation}
m(c,T)=\left\{\begin{matrix}
 min(\frac{T}{T_k} \tau ,\tau ) &  c\in S_n ,\\
 1 & c \in S_c ,
\end{matrix}\right.
\end{equation}
Specifically, $m(c)$ should be a non-increasing function by $T$, which means that the penalty factor can not be boundless as the $T$ increasing. $T_k$ depends on the noise rate. Then, for each layer $l$, we compute a weighted sum to mitigate the negative impact from the noisy clients while improving the generalization of the global model as : 
\begin{equation}
w^{c}_l = \frac{m(c,T) D_l^c}{\sum_{c=1}^{C} m(c,T) D_l^c},
\end{equation}
Finally, we perform the layer-wise weighted aggregation of the local models and update the global model as:
\begin{equation}
\Theta^G_{t+1} \longleftarrow \sum_{l=1}^{L}\sum_{c=1}^{C} w_l^{c} \Theta^c_{l_t},
\label{eq_aggre}
\end{equation}
where $\Theta^G_{t+1}$ represents the parameters of the aggregated global model and $\Theta^c_{l_t}$ denotes the parameters of layer $l$  received from client $c$ after the global training round $t$.
In the aggregation process, the weight of a local model is inversely proportional to the distance between the local model and the updated global model. Thus, as the training proceeds, the global model will update in the direction that the clean clients guide it to.


\subsection{Label Correction}
To fully utilize the local data, we perform the label correction on the noisy clients' local data. We follow the two stages training procedure like \cite{huang2019o2u}. Suppose the global model converges after $T_{corr}$ round training, we apply the label correction on the noisy clients $S_n$. To ensure the label correction performs on the real noisy clients, we extend the validation time interval at different training iterations to identify the real noisy clients. At each global training round, the clients are divided into noisy clients and clean clients based on the method we proposed in Sec.~\ref{noisydetection}. Before $T_{corr}$-1 round, the server stores each round's results of noisy client detection. The noisy clients $S_{corr}$need to conduct label corrections are determined as:
\begin{equation}
    S_{corr}= (\sum_{t=1}^{T_{corr}}\vmathbb{1}(c \in S_n^t)) > \alpha T_{corr}.
\end{equation}
Then, each noisy client $c \in S_{corr}$ performs the label correction of all local data by using the predicted label from the global model $\Theta_G$ to generate a new pseudo label. To avoid over-correction, we only relabel the instances with conﬁdence exceeding the threshold $\eta$. Thus, for each noisy client $c \in S_{corr}$, the subset $\widetilde{D}_c$ of relabel samples is given by:
\begin{equation}
 \widetilde{D}_c=\{(x^c,\hat{y}^c)|\hat{y}^c=\max(f(x^c;\Theta_t^G)>\eta)\}.   
\end{equation}
After performing the label correction on the noisy clients $S_{corr}$, the new relabel subset $\widetilde{D}$ is used for local model training.

\section{Experiments} 
\label{sec_exp}


\subsection{Experimental Setup}
\noindent{\textbf{Datasets and Models.}} We conduct experiments based on three datasets: 1) MNIST~\cite{mnist}, 2) FASIONMNIST~\cite{fasionmnist} 3) CIFAR10~\cite{cifar10}, which are widely used for the evaluation of noisy labels in previous works~\cite{CoTeaching}. 
In our experiment, we evaluate the performance of \ourmodel~on the following two widely-used deep learning models:  Lenet5~\cite{Lenet5} and VGG16~\cite{VGG}. 

\noindent{\textbf{Baseline.}} We evaluate the effectiveness of \ourmodel~with the following three FL baselines.
\textbf{FedAvg~\cite{FedAvg_McMahan2017}} conducts local training on the participating clients and then performs model aggregation on the central server only considering the training data size on each corresponding client. The more training data a client contains, the higher weight it has in the aggregation process. 
\textbf{Trimmed mean~\cite{trimmedmean}} aggregates each model parameter independently. In particular, for the $c$-th model parameter in the collaborative model, the central server sorts the $c$-th parameters of the local models received from the N participating clients. After that, it removes the largest and smallest $K$ percentage of them and computes the mean of the remaining parameters as the $c$-th parameter of the global model. 
\textbf{FedProx~\cite{FedProx_Li2018}} adds a proximal term $\frac{\mu}{2} || \Theta_c-\Theta_{G}^t||^2$ on clients' local optimization objective to mitigate the negative effect caused by local data heterogeneity. The $\mu$ is the hyperparameter that limits the optimation divergence between the local and global models. We set the $\mu=0.01$, which is the same as \cite{FedProx_Li2018}. The proximal term can reduce the variance of local updates since it encourages the local update to be closer to the global model.

\noindent{\textbf{Noise Distribution.}} In the experiment, we evaluate the performance of \ourmodel~under different noisy data distributions. Specifically, the following two common distributions are adopted to characterize the noise scenario across different clients: 1) Bernoulli distribution and 2) truncated Gaussian distribution.

\noindent{\textbf{Data Distribution.}} To simulate the real-world scenario, we consider both IID and Non-IID data partitions in FL. For the IID data partitions, we uniformly divide the whole dataset to $C$ clients. For the Non-IID data partitions, we consider two heterogeneous data scenarios across the clients: quantity and class distribution skewness. To simulate the class distribution skewness, we follow \cite{xu2022fedcorr} to perform the data partition. Specifically, for a certain class $j$, it is sampled from the Bernoulli distribution with a fixed probability $p$ to generate the class distribution among all the clients. This class distribution indicates whether the local dataset of client $c$ contains class $j$. Then the number of class $c$ of training samples of class $j$ in client $c$ is sampled from the symmetric Dirichlet distribution with the parameter $\alpha_{DIR}$. To simulate the quantity skewness, we sample the size of training data among the clients from a lognormal distribution. For all the datasets, we set the standard deviation of sample size as 0.3.

\noindent{\textbf{Implementation Detail.}} For all the experiments, we set the number of clients as $N=20$, the total global rounds as 150, and the $T_{corr}=60$. We use SGD as the local training optimizer with a learning rate of 0.01, with a local batch size of 60 and local epoch of 10 for all datasets. With the implementation setting of \ourmodel, we set the hyperparameter $\alpha$ with 0.6 and $\tau$ with 50 for all experiments. For the noise scenario with Bernoulli distribution for all Gaussian distribution for all the datasets, we set $\beta$ as 0.6 for all the experiments.

\begin{table}[!t]
	\begin{center}  
		\renewcommand\arraystretch{1.3}
		\caption{Accuracy of different schemes in various scenarios when each client has the \textbf{same amount} of local training data. We report the average accuracy over the last 10 rounds.
		(S1-Bernoulli distribution with $p=0.6$; S2-Bernoulli distribution with $p=0.7$; S3-Bernoulli distribution with $p=0.8$;  S4-Truncated Gaussian Distribution with $(\mu=0.4, \sigma = 0.45)$; S5-Truncated Gaussian Distribution with $(\mu=0.3, \sigma = 0.45)$; S6-Truncated Gaussian Distribution with $(\mu=0.3, \sigma = 0.4)$).\label{sumuniform}} 
		\resizebox{1.0\linewidth}{!}{
			\begin{tabular}{|c|c|c|c|c|c|c|} 
				\hline
				& \textbf{S1} & \textbf{S2} & \textbf{S3} & \textbf{S4} & \textbf{S5} & \textbf{S6} \\\hline
				\multicolumn{7}{|l|}{\textbf{Lenet5 on MNIST}} \\\hline
				Ours & \textbf{98.87\%} & \textbf{98.84\%} & \textbf{99.12\%} & \textbf{97.24\%} & \textbf{97.65\%} & \textbf{97.04\%} \\\hline
				FedAvg & 87.64\% & 93.96\% & 88.00\% & 86.63\% & 88.78\% & 89.08\% \\\hline
				Trimm & 82.62\% & 97.59\% & 98.72\% & 91.35\% & 95.01\% & 94.46\% \\\hline
				FedProx & 83.24\% & 88.27\% & 96.12\% & 85.73\% & 88.90\% & 89.79\% \\\hline
                
				\multicolumn{7}{|l|}{\textbf{Lenet5 on Fashion MNIST}} \\\hline
				Ours & \textbf{87.48\%} & \textbf{88.53\%} & \textbf{88.85\%} & \textbf{85.29\%} & \textbf{86.23\%} & \textbf{86.73\%} \\\hline
				FedAvg & 67.61\% & 81.30\% & 75.93\% & 75.23\% & 78.32\% & 79.29\% \\\hline
				Trimm & 60.55\% & 76.55\% & 85.93\% & 78.14\% & 82.54\% & 81.79\% \\\hline
				FedProx & 65.91\% & 76.29\% & 76.60\% & 79.02\% & 81.23\% & 81.22\% \\\hline
                
				\multicolumn{7}{|l|}{\textbf{VGG16 on CIFAR10}} \\\hline
				Ours & \textbf{72.87\%} & \textbf{77.97\%} & \textbf{78.28\%} & \textbf{62.88\%} & \textbf{67.37\%} & \textbf{66.50\%} \\\hline
				FedAvg & 42.82\% & 51.16\% & 58.19\% & 51.17\% & 54.99\% & 57.01\% \\\hline
				Trimm & 41.85\% & 64.87\% & 73.09\% & 54.25\% & 62.45\% & 60.82\%\\\hline
				FedProx & 43.50\% & 48.84\% & 56.41\% & 50.65\% & 55.88\% & 55.95\% \\\hline
               
			\end{tabular}
		}
			\label{sumrandom}
	\end{center} 
\vspace{-0.1 in }
\end{table}

\begin{figure*}[!t] 
	\centering
	\includegraphics[width=1\linewidth]{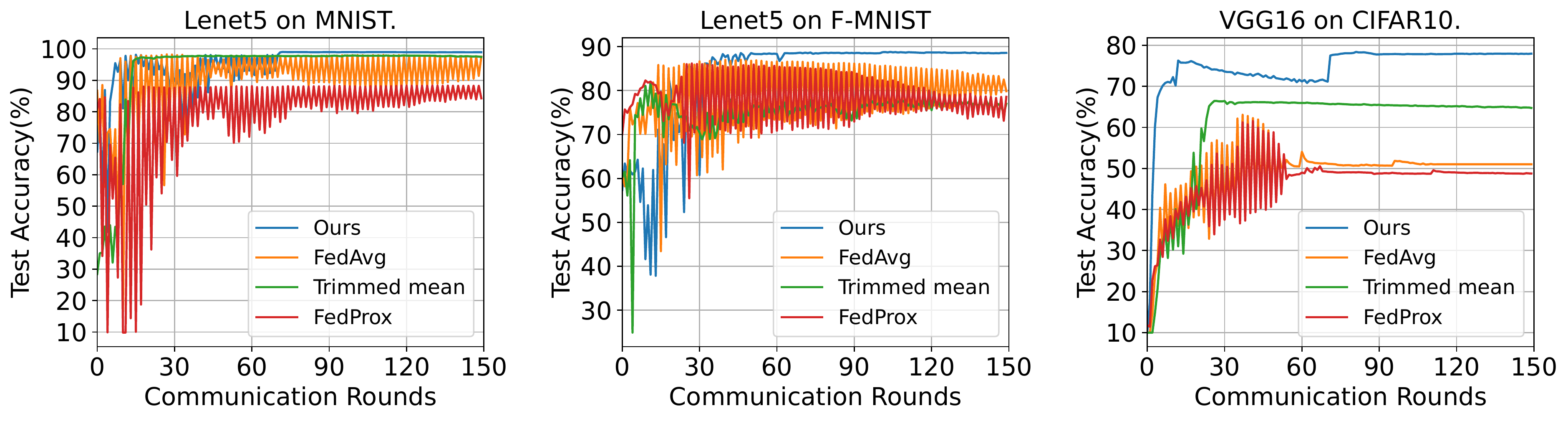}
	\caption{Comparison of different schemes when the noisy data across different clients follows a Bernoulli distribution with p = 0.7. The global model trained by baseline methods show the unstable prediction on test dataset, which demonstrates the global model is difficult to converge while existing noisy clients. }
	\label{uniformresult1}
\end{figure*}

\begin{table}[!t]
  \begin{center} 
  \renewcommand\arraystretch{1.3}
    \caption{Accuracy of different schemes in various scenarios when the participating clients have \textbf{different amounts} of training data. We report the average accuracy over the last 10 rounds. (S1-Bernoulli distribution with $p=0.6$; S2-Bernoulli distribution with $p=0.7$; S3-Bernoulli distribution with $p=0.8$;  S4-Truncated Gaussian Distribution with $(\mu=0.4, \sigma = 0.45)$; S5-Truncated Gaussian Distribution with $(\mu=0.3, \sigma = 0.45)$; S6-Truncated Gaussian Distribution with $(\mu=0.3, \sigma = 0.4)$).
    } 
    \scriptsize
    \resizebox{1.0\linewidth}{!}{
    \begin{tabular}{|c|c|c|c|c|c|c|} 
    \hline
         & \textbf{S1} & \textbf{S2} & \textbf{S3} & \textbf{S4} & \textbf{S5} & \textbf{S6} \\\hline
        \multicolumn{7}{|l|}{\textbf{Lenet5 on MNIST}} \\ \hline
        Ours & \textbf{98.72\%} & \textbf{98.97\%} & \textbf{98.58\%} & \textbf{95.54\%} & \textbf{97.20\%} & \textbf{96.88\%} \\\hline
        FedAvg & 82.75\% & 83.76\% & 97.06\% & 84.18\% & 89.86\% & 90.47\% \\\hline
        Trimm & 63.32\% & 97.35\% & 98.48\% & 92.01\% & 94.92\% & 94.78\% \\\hline
        FedProx & 83.51\% & 95.97\% & 96.65\% & 82.87\% & 89.57\% & 89.71\% \\\hline
        
        \multicolumn{7}{|l|}{\textbf{Lenet5 on Fashion MNIST}} \\\hline
        Ours & \textbf{84.34\%} & \textbf{87.71\%} & \textbf{88.36\%} & \textbf{82.02\%} & \textbf{84.90\%} & \textbf{85.70\%} \\\hline
        FedAvg & 68.82\% & 75.85\% & 86.54\% & 75.44\% & 80.00\% & 80.38\% \\\hline
        Trimm & 65.61\% & 77.89\% & 86.18\% & 79.69\% & 82.50\% & 82.17\% \\\hline
        FedProx & 73.76\% & 85.22\% & 85.94\% & 73.44\% & 78.83\% & 78.79\% \\\hline
        
        \multicolumn{7}{|l|}{\textbf{VGG16 on CIFAR10}} \\\hline
        Ours & \textbf{53.54\%} & \textbf{68.56\%} & \textbf{79.71\%} & \textbf{58.65\%} & \textbf{62.91\%} & \textbf{65.24\%} \\\hline
        FedAvg & 44.87\% & 55.65\% & 64.74\% & 45.05\% & 56.82\% & 58.51\% \\\hline
        Trimm & 43.57\% & 54.84\% & 69.58\% & 54.10\% & 61.20\% & 62.49\%\\\hline
        FedProx & 43.56\% & 52.90\% & 54.99\% & 55.65\% & 58.21\% & 54.50\% \\\hline
        
        \end{tabular}
    }
    \label{sumunbalanced}
  \end{center} 
  \vspace{-0.1 in }
\end{table}

\subsection{Evaluation with Heterogeneous Noise Distributions}
\label{exp_Uniform}
\begin{figure*}[!t] 
	\centering
	\includegraphics[width=1\linewidth]{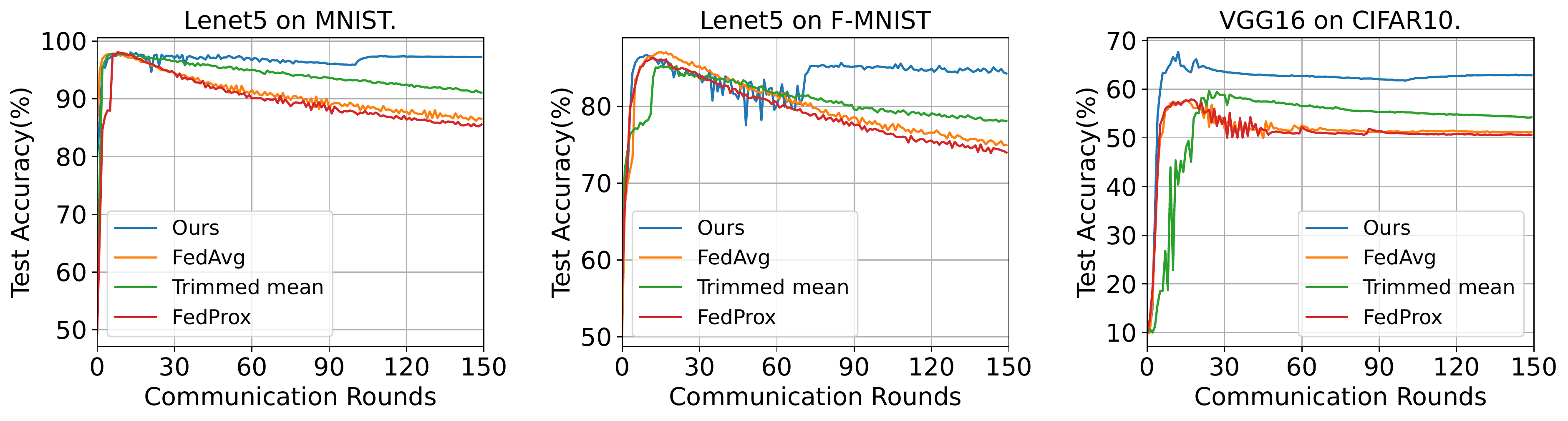}
	\caption{Comparison of different schemes when the noisy data across different clients follows truncated Gaussian distribution with ($\mu$= 0.4, $\sigma$ = 0.45).}
	\label{gaussian}
\end{figure*}

In this section, we first evaluate the effectiveness of our \ourmodel~on two noise scenarios proposed in \cref{sec_noisy}, when the training data are uniformly distributed among the participating clients.
Tab.~\ref{sumrandom} shows the evaluation results of different schemes when the noisy data follows the Bernoulli distribution and truncated Gaussian distribution. The experimental results demonstrate that \ourmodel~outperforms the other baselines in various noise distributions. For Bernoulli distribution, the Trimmed mean and FedProx cannot address the noise clients problem as their average test accuracy drops significantly when increasing the number of noisy clients, by 24.23\% for the Trimmed mean and 12.16\% for the FedProx, averagely. For \ourmodel, it only drops 2.34\%. Significantly, \ourmodel~outperforms other baselines by at least 11.23\% in MNIST, 19.87\% in Fashion MNIST, and 29.37\% in CIFAR10 in the S1 setting (60\% of clients with noisy labels data). These results point out that \ourmodel~is robust to noisy clients, especially in the high ratio of noisy clients. For the truncated Gaussian distribution, the corresponding results show that \ourmodel~also can effectively mitigate the impact of noisy clients which follows a truncated Gaussian distribution, which improves the FedAvg at least by 7.96\% in MNIST,  7.44\% in Fashion MNIST, and 9.49\% in CIFAR10. 
Figs.~\ref{uniformresult1} and \ref{gaussian} show the global model's test accuracy of different schemes when the noisy data follows a Bernoulli distribution with $p=0.7$ across different clients and truncated Gaussian distribution with $(\mu=0.4, \sigma = 0.45)$, respectively. 
Specifically, for the Bernoulli distribution, we can see that the global model trained by FedAvg and FedProx can not converge because their test accuracy curves are unstable at the end of training. This is because the noisy clients guide the update of the collaborative model in a divergent direction during the model aggregation process. Although the FedProx adds the extra proximal term to reduce the variance of local updates, it can not effectively mitigate the negative effect from the noisy data. The Trimmed mean performs better by removing a specific percentage of the smallest and largest parameters from the collected local models. However, \ourmodel~achieves the prominent performance advantage among all the baselines for the following reasons. 
During the training process, \ourmodel~can prevent the misleading from the noisy clients' model by correctly identifying the noisy clients from clean clients. The reliability score can effectively evaluate each client by measuring the quality of the local training data and model. Thus, correctly detecting the noisy clients is the first key success of \ourmodel. Then, our robust layer-wise aggregation can reduce the impact of the noisy clients and guide the global model's updates with clean clients' model, increasing the model convergence rate. Finally, \ourmodel~dynamically performs label correction on the noisy clients, which improves the generalization ability of the global model.

\subsection{Evaluation with Heterogeneous Data Distributions}
In this section, we evaluate the performance of different schemes when the participating clients with heterogeneous local training data while exiting the noisy label data.  
Tab.~\ref{sumrandom} shows the evaluation results of different schemes when the noisy data follows the Bernoulli distribution and truncated Gaussian distribution while the number of local training data across clients is unbalanced. \ourmodel~consistently outperforms other methods, over up to 35\%. Tab.~\ref{sumrandom} shows the evaluation results of different schemes when the noisy data follows Bernoulli distribution and truncated Gaussian distribution while the distribution of clients' data is heterogeneous. \ourmodel~is effective to simultaneously tackle the client with noisy labels data and heterogeneous data, by over other methods up to 26\%.
Compared with the data partition with the IID setting, all the baselines have performance degradation. On the contrary, \ourmodel~is still effective in different noise scenarios with data heterogeneity. The challenge of tackling noisy clients with data heterogeneity is that we cannot quantify the impact of the data heterogeneity and noisy clients; both of them direct the collaborative model in a divergent direction. To address this challenge, \ourmodel~first identifies the noisy clients from the clean clients by measuring the client's data quality and model divergence, which can help the global model update with a correct direction with SGD. Then the robust layer-wise aggregation can aggregate all the clients' models in layer-quantify to gain a more robust global model. Therefore, \ourmodel~is effective for tackling noisy clients with data heterogeneity. 

\subsection{Ablation Study}
In this section, to study the effectiveness of \ourmodel, we conduct the ablation study of \ourmodel~on the CIFAR-10 in two noise scenarios proposed in \cref{sec_noisy} with different data partitions. Tab.~\ref{ablationstudy} shows the corresponding result. The result shows that adding the penalty for noisy clients in layer-wise model aggregation has a critical contribution to \ourmodel. Without the penalty for noisy clients' local models, the performance of global models can decrease at most 26.58\%, and 25.32\% in IID, Non-IID data partitions, respectively. It represents that adding the penalty for the noisy clients in layer-wise model aggregation can prevent the negative effect from the noisy clients during the model aggregation. Thus, it can improve the effectiveness and robustness of global training in federated noisy client learning. For the label correction, it can improve the global model performance up to 13\% in serious noise settings (30\% of clients have data with noisy labels). As for the noisy client's detection, precisely measuring the reliability score of each client is the key to success in effectively detecting noisy clients. Thus, we removed two corresponding criteria in Eqs.~(\ref{modeldistance}) and (\ref{CEloss}) in reliability score, respectively. We can observe that both criteria contribute to reliability score, which improves global model's performance 6\% by on average. 

\begin{table}[!t]
\begin{center}
\renewcommand\arraystretch{1.3}
 \caption{Accuracy of non-iid data. We report the average accuracy over the last 10 rounds. 	(S1-Bernoulli distribution with $p=0.7$; S2-Truncated Gaussian Distribution with $(\mu=0.4, \sigma = 0.45)$).} 
    \scriptsize
    \resizebox{1.0\linewidth}{!}{
\begin{tabular}{|l|cc|cc|cc|}
\hline
Dataset/Model & \multicolumn{2}{c|}{Lenet5 on Mnist} & \multicolumn{2}{c|}{Lenet5 on F-Mnist} & \multicolumn{2}{c|}{VGG16 on CIFAR10} \\ \hline
Noise Setting & \multicolumn{1}{c|}{S1}    & S2    & \multicolumn{1}{c|}{S1}    & S2    & \multicolumn{1}{c|}{S1}    & S2    \\ \hline
Ours          & \multicolumn{1}{c|}{\textbf{98.79\%}}     &    \textbf{94.15\%}  & \multicolumn{1}{c|}{\textbf{88.19\%}} & \textbf{83.56\%} & \multicolumn{1}{c|}{\textbf{77.19\%}} & \textbf{57.16\%} \\ \hline
FedAvg        & \multicolumn{1}{c|}{94.28\%} & 85.45\% & \multicolumn{1}{c|}{75.48\%} & 75.59\% & \multicolumn{1}{c|}{53.50\%} & 52.82\% \\ \hline
Trimm         & \multicolumn{1}{c|}{97.81\%} & 92.39\% & \multicolumn{1}{c|}{75.32\%} & 78.91\% & \multicolumn{1}{c|}{60.87\%} & 52.25\% \\ \hline
FedProx       & \multicolumn{1}{c|}{90.68\%} & 85.95\% & \multicolumn{1}{c|}{68.94\%} & 73.60\% & \multicolumn{1}{c|}{50.68\%}  & 53.99\% \\ \hline
\end{tabular}
}
\end{center}
\end{table}

\begin{table}[]
\centering
\renewcommand\arraystretch{1.4}
 \caption{Ablation study. We report the average accuracy over the last 10 rounds on CIFAR-10. (S1-Bernoulli distribution with $p=0.7$; S2-Bernoulli distribution with $p=0.8$; S3-Truncated Gaussian Distribution with $(\mu=0.4, \sigma = 0.45)$); S4-Truncated Gaussian Distribution with $(\mu=0.3, \sigma = 0.4)$).} \label{ablationstudy}
\scriptsize
\resizebox{1.0\linewidth}{!}{
\begin{tabular}{|l|cccc|cccc|}
\hline
Data partition     & \multicolumn{4}{c|}{IID}    & \multicolumn{4}{c|}{Non-IID}  \\ \hline
Noise Setting      & S1  & S2   & S3 & S4   & S1& S2  & S3  & S4   \\ \hline
w/o label correction            &    64.89\%                  &  76.31\%             &    55.26\%                  &    59.10\%                    &   64.97\%         &     77.13            &   57.02\%                   &    61.88\%                  \\
w/o penalty      &  51.12\%       &    56.60\%        &    47.50\%      &     54.40\%    &  51.87\%       &   58.99\%        &    51.94\%      &        56.66\%              \\
w/o criteria Eq.~(\ref{modeldistance})  &     74.27\%        &   77.03\%        &     56.98\%      &   59.34\%       &    68.95\%                  &      76.70\%   &  57.04\%         &   61.75\%         \\
w/o criteria Eq.~(\ref{CEloss})   &     70.89\%  &  77.15\%&  46.84\%            &  54.67\%      &      55.55\%         &    61.37\%         &     51.21\%          &           57.39\%     \\ \hline
Ours                            & \multicolumn{1}{l}{77.97\%} & \multicolumn{1}{l}{78.28\%} & \multicolumn{1}{l}{58.65\%} & \multicolumn{1}{l|}{65.24\%} & \multicolumn{1}{l}{77.19\%} & \multicolumn{1}{l}{78.59\%} & \multicolumn{1}{l}{57.16\%} & \multicolumn{1}{l|}{61.90\%} \\ \hline
\end{tabular}
}
\end{table}

\section{Conclusion}  
\label{sec_conclusion}
In this paper,  we proposed two kinds of heterogeneous noisy label scenarios in FL to precisely model the noise level distribution across the clients. 
To study the impact of noisy clients on federated learning, we first investigate critical issue caused by noisy clients in FL and quantify the negative impact of the noisy clients in terms of the representations learned by different layers. We have shown that the noisy client could negatively impact the convergence, feature extraction ability, and performance of the aggregated global model in FL. 
Moreover, in order to effectively conduct FL with noisy clients, we have proposed a simple yet effective learning framework, Federated Noisy Client Learning (\ourmodel), which first identifies the noisy clients through well estimating the data quality and model divergence. Then robust layer-wise aggregation is proposed to adaptively aggregate the local models of each client to deal with the data heterogeneity caused by the noisy clients. We further perform the label correction on the noisy clients to improve the generalization of the global model.
Experimental results have shown that our \ourmodel~can effectively boost the performance of different deep networks with noisy clients in homogeneous and heterogeneous FL.
%

\bibliographystyle{IEEEtran}
	\bibliography{NoisyFL}

\end{document}